\documentclass[a4paper,fleqn]{cas-dc}

\usepackage[numbers,sort,compress]{natbib}

\usepackage{algorithm}
\usepackage{algorithmic}

\usepackage{hyperref}
\usepackage{caption}

\usepackage{graphicx}
\usepackage{multirow}
\usepackage{makecell}
\usepackage{booktabs}
\usepackage{bbding}
\usepackage{algorithm}
\usepackage{algorithmic}
\usepackage{amsmath}

\makeatletter
\newenvironment{breakablealgorithm}
  {
    \begin{center}
      \refstepcounter{algorithm}
      \hrule height.8pt depth0pt \kern2pt
      \parskip 0pt
      \renewcommand{\caption}[2][\relax]{
        {\raggedright\textbf{\fname@algorithm~\thealgorithm} ##2\par}%
        \ifx\relax##1\relax 
          \addcontentsline{loa}{algorithm}{\protect\numberline{\thealgorithm}##2}%
        \else 
          \addcontentsline{loa}{algorithm}{\protect\numberline{\thealgorithm}##1}%
        \fi
        \kern2pt\hrule\kern2pt
     }
  }
  {
     \kern2pt\hrule\relax
   \end{center}
  }
\makeatother

\def\tsc#1{\csdef{#1}{\textsc{\lowercase{#1}}\xspace}}
\tsc{WGM}
\tsc{QE}
\tsc{EP}
\tsc{PMS}
\tsc{BEC}
\tsc{DE}

\begin{document}
\captionsetup[figure]{labelfont={bf},labelformat={default},labelsep=period,name={Fig.}}
\let\WriteBookmarks\relax
\def\floatpagepagefraction{1}
\def\textpagefraction{.001}
\let\printorcid\relax

\shorttitle{MMWOZ: Building Multimodal Agent for Task-oriented Dialogue}

\shortauthors{Pu-Hai Yang et~al.}

\title [mode = title]{MMWOZ: Building Multimodal Agent for Task-oriented Dialogue}                      



%

\author[1]{Pu-Hai Yang}
\cormark[1]
\ead{phyang@ahu.edu.cn}
\credit{Conceptualization, Methodology, Software, Validation, Formal analysis, Investigation, Data curation, Writing - original draft}
\address[1]{School of Artificial Intelligence, Anhui University, Hefei, China}

\author[2]{Heyan Huang}
\ead{hhy63@bit.edu.cn}
\credit{Writing - review \& editing, Resources, Supervision, Project administration, Funding acquisition}
\address[2]{School of Computer Science and Technology, Beijing Institute of Technology, Beijing, China}

\author[2]{Heng-Da Xu}
\ead{xuhengda@bit.edu.cn}
\credit{Revise the paper, Writing - review \& editing}

\author[2]{Fanshu Sun}
\ead{sunfs@bit.edu.cn}
\credit{Revise the paper, Writing - review \& editing}

\author[2]{Xian-Ling Mao}
\ead{maoxl@bit.edu.cn}
\credit{Write part of the paper and revise the paper, Supervision, Formal analysis, Writing - review \& editing}

\author[1]{Chaoxu Mu}
\ead{cxmu@tju.edu.cn}
\credit{Writing - review \& editing, Resources, Supervision, Project administration}

\cortext[cor1]{Corresponding author}

\begin{abstract}
Task-oriented dialogue systems have garnered significant attention due to their conversational ability to accomplish goals, such as booking airline tickets for users. Traditionally, task-oriented dialogue systems are conceptualized as intelligent agents that interact with users using natural language and have access to customized back-end APIs. However, in real-world scenarios, the widespread presence of front-end Graphical User Interfaces (GUIs) and the absence of customized back-end APIs create a significant gap for traditional task-oriented dialogue systems in practical applications. In this paper, to bridge the gap, we collect MMWOZ, a new multimodal dialogue dataset that is extended from MultiWOZ 2.3 dataset. Specifically, we begin by developing a web-style GUI to serve as the front-end. Next, we devise an automated script to convert the dialogue states and system actions from the original dataset into operation instructions for the GUI. Lastly, we collect snapshots of the web pages along with their corresponding operation instructions. In addition, we propose a novel multimodal model called \textbf{MATE} (\textbf{M}ultimodal \textbf{A}gent for \textbf{T}ask-ori\textbf{E}nted dialogue) as the baseline model for the MMWOZ dataset. Furthermore, we conduct comprehensive experimental analysis using MATE to investigate the construction of a practical multimodal agent for task-oriented dialogue.
\end{abstract}


\maketitle

\section{Introduction}

Task-oriented dialogue systems aim to accomplish various user goals through natural language communication, which often involve complexity and require multiple dialogue turns to complete \cite{valizadeh-parde-2022-ai,algherairy2023review, ni2023recent}. For instance, when assisting users in booking air tickets, a task-oriented dialogue system engages in a conversation to gather information such as the departure place, destination, and departure time. Once sufficient information is obtained, the system automatically handles the booking process. The convenience offered by this natural language interaction has led to a growing interest in task-oriented dialogue systems in recent years \cite{yang2021ubar, su-etal-2022-multi, hudecek-dusek-2023-large}.

\begin{figure}
    \centering
    \includegraphics[width=\linewidth]{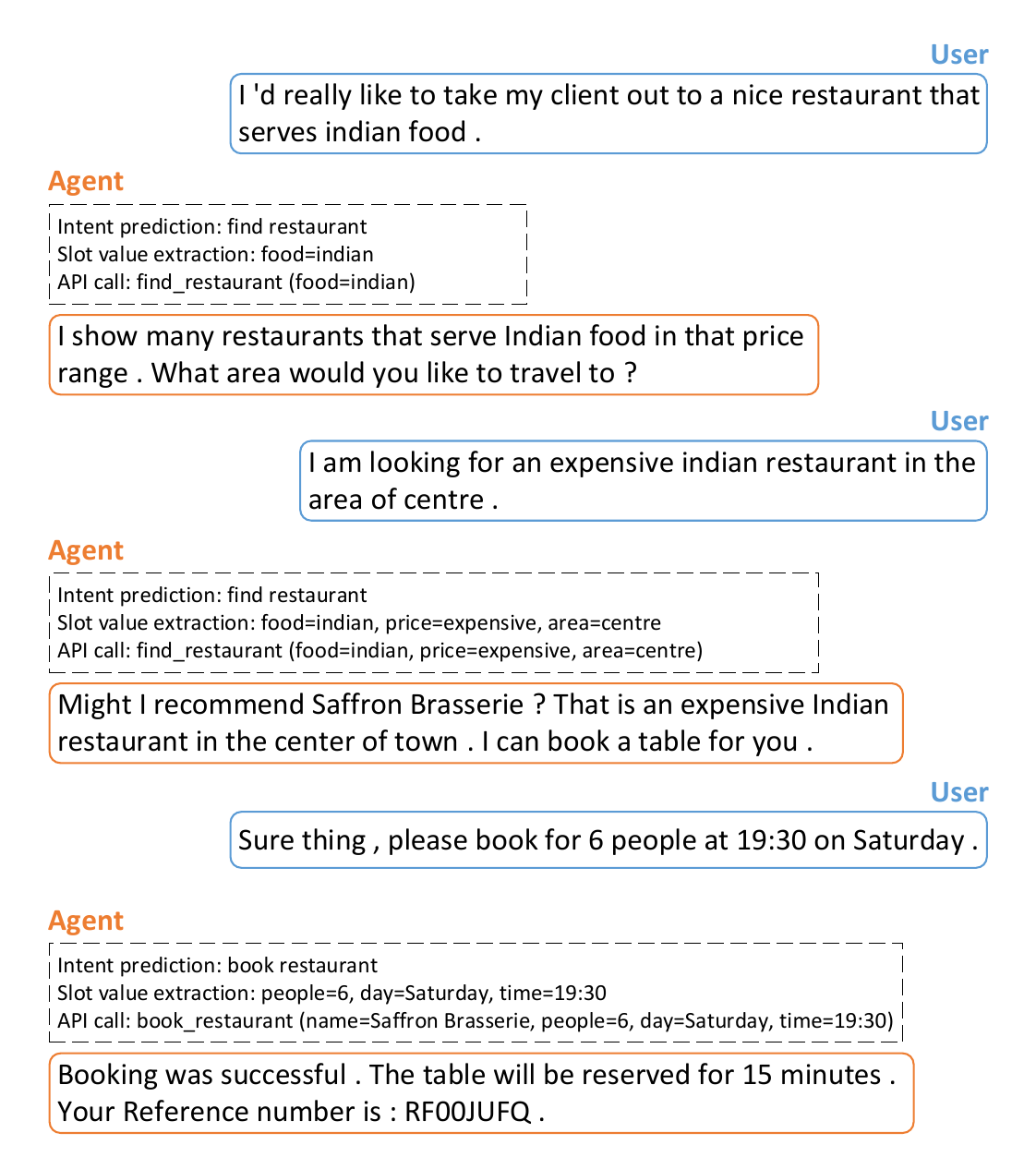}
    \caption{An example of an agent interacting with a user in a traditional task-oriented dialogue system (From dialogue "MUL0001" in the MultiWOZ 2.3 dataset).}
    \label{TODS}
\end{figure}

\begin{figure*}
    \centering
    \includegraphics[width=\linewidth]{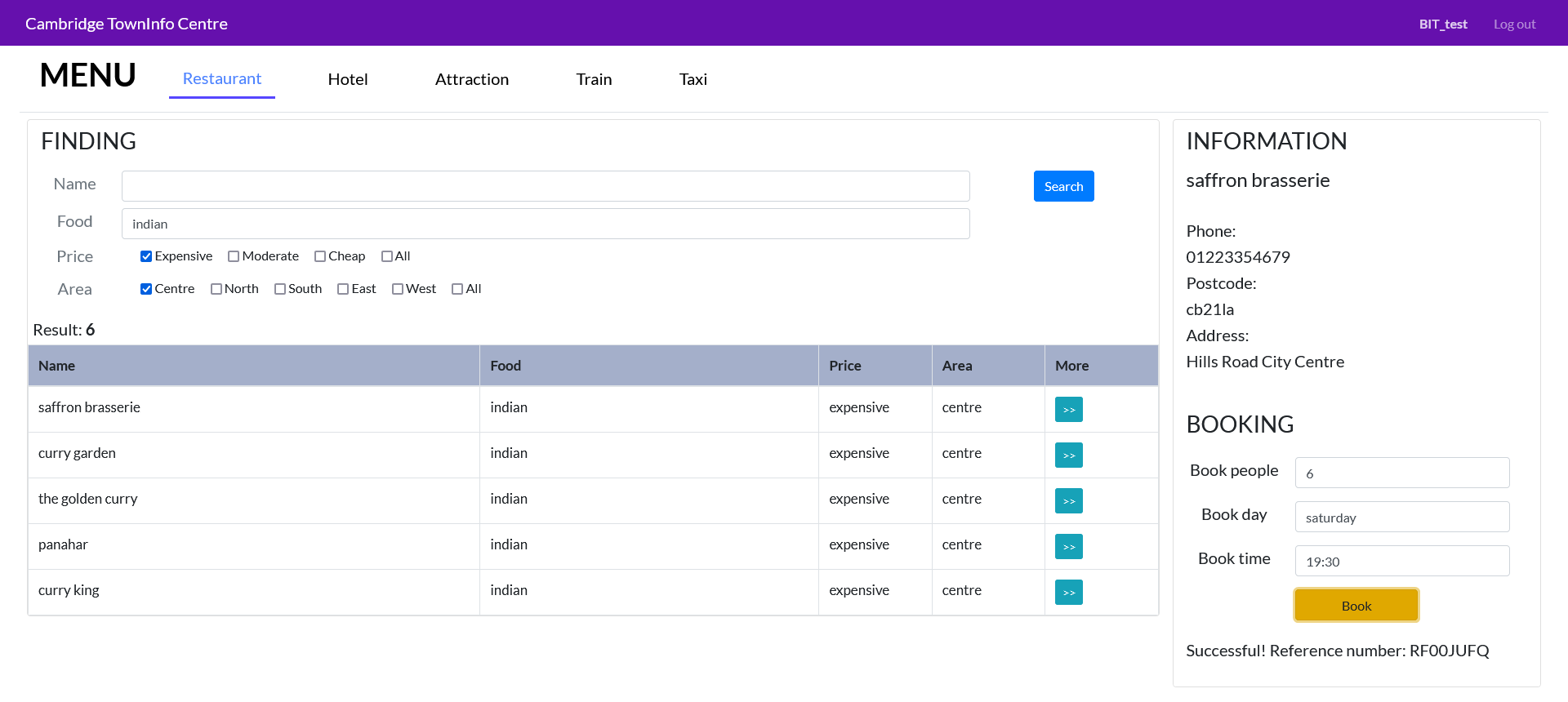}
    \caption{An example of using a web-style GUI to find and book a restaurant. The agent find 6 expensive Indian restaurants in the town centre and eventually book the user a table for 6 people at saffron brasserie for 19:30 on Saturday (Snapshot of the web page obtained by the agent after executing GUI operation instructions in the last turn in Figure \ref{TODS}).}
    \label{GUI}
\end{figure*}

Traditionally, task-oriented dialogue systems are generally modeled as intelligent agents that have access to back-end APIs to acquire knowledge in a database \cite{budzianowski-etal-2018-multiwoz, zhu-etal-2020-crosswoz, rastogi2020towards}, thereby using this knowledge to help users complete various tasks. These agents follow a pipeline process in the dialogue with users: predict the user's intention, extract slot values in the user's utterance, call API to access the database and response to the user \cite{ijcai2021p622, jacqmin-etal-2022-follow, kwan2023survey, wu-etal-2023-diacttod, hosseini2020simple, peng-etal-2021-soloist}. For example, as shown in Figure \ref{TODS}, when a user desires to book a restaurant, the agent engages in a dialogue where, in the first 4 turns, the user seeks a restaurant meeting specific requirements, prompting the agent to call the "find\_restaurant" API. In the last 2 turns, the user provides detailed reservation information, leading to the agent calling the "book\_restaurant" API.

However, in real-world scenarios, the availability of customized APIs for building practical task-oriented dialogue systems is limited, primarily due to two reasons. Firstly, commercial websites and applications often refrain from disclosing their internal API documentation to protect data security and maintain a competitive advantage. Secondly, task-specific APIs that are comprehensible and easily integrated into dialogue systems can be expensive. In contrast, graphical user interfaces (GUIs) have become the predominant means of interacting with websites and apps, which offer a user-friendly and information-rich interaction mode. Consequently, the traditional API-based approach for task-oriented dialogue systems faces challenges in widespread adoption for practical applications.

To bridge the gap between task-oriented dialogue system in ideal scenario and practical application, in this paper, we propose the MMWOZ dataset, which is a multimodal task-oriented dialogue dataset containing interactions with a simplified GUI, and is extended from MultiWOZ 2.3 \cite{han2021multiwoz}, an upgraded version of MultiWOZ dataset \cite{budzianowski-etal-2018-multiwoz}. To build the MMWOZ dataset, specifically, we first design a web-style GUI and make it accessible to the database content provided in the MultiWOZ 2.3 dataset. Then, relying on a carefully crafted automated script, we convert the dialogue state and system action annotations from the MultiWOZ 2.3 dataset into a series of operation instructions for the GUI. Finally, along the execution sequence of instructions, we collect the operation instructions and corresponding snapshots of the web page. Figure \ref{GUI} shows the snapshot of the web page after an agent successfully searched and booked the restaurant for a user in our MMWOZ dataset, different from the traditional API-based task-oriented dialogue system, the agent needs to understand the user's requirements and convert them into operation instructions for the GUI, and then reply to the user according to the results in the snapshot of the web page.

Furthermore, we propose MATE, a novel multimodal model, as a baseline on the MMWOZ dataset. MATE leverages dialogue history, action log, and a web page snapshot to generate operation instructions for the GUI or a natural language response to the user. Additionally, we perform an in-depth analysis of MATE's performance on the MMWOZ dataset, aiming to uncover effective strategies for constructing practical multimodal agents for task-oriented dialogues. To summarize, this paper offers the following key contributions: (1) We collect MMWOZ, a multimodal task-oriented dialogue dataset designed for interactions with a web-style GUI, which sets a new benchmark in comparison to traditional API-based task-oriented dialogues. (2) We propose MATE, a novel multimodal task-oriented dialogue model, which can control web-style GUI by generating operation instructions and respond to users by utilizing information from the snapshot of the web page. (3) We perform comprehensive experimental analysis on the MMWOZ dataset to investigate the key factors that need to be considered when constructing a multimodal task-oriented dialogue agent with the ability to control the GUI.

For the organization of the rest of this paper: Section 2 provides a detailed introduction to the work related to task-oriented dialogue, multimodal dialogue, and GUI navigation. Section 3 elaborates the collection method and statistical information of the dataset in this paper. The architecture of the proposed model is described in Section 4. Section 5 introduces the setting of training parameters and evaluation metrics. Section 6 presents results, experimental analysis and cases, followed by a conclusion in section 7.

\section{Related Work}
\subsection{Task-oriented Dialogue}

Due to its ability to help people complete diverse and complex tasks, task-oriented dialogue system (TODS) has received a lot of attention in research and application. Traditionally, task-oriented dialogue system is generally composed of four modules: (1) Natural Language Understanding (NLU): This component analyzes user utterances to extract intent and slot values, identifying what the user wants to achieve and the relevant parameters (e.g., location, time) \cite{goo-etal-2018-slot,zhu-etal-2020-multitask}. (2) Dialogue State Tracking (DST): It maintains a representation of the conversation context, keeping track of the current state of the dialogue and user goals \cite{wu-etal-2019-transferable,kim-etal-2020-efficient}. (3) Dialogue Policy Learning (DPL): Decides on the next system action based on the dialogue state, such as requesting more information, offering options, or executing a task \cite{takanobu-etal-2020-multi,zhao-etal-2024-bootstrapped}. (4) Natural Language Generation (NLG): Converts system actions into natural language responses that are coherent and relevant to the ongoing dialogue \cite{wang-etal-2020-multi-domain,du-etal-2024-rewarding}. In recent years, with the rapid development of pre-trained language models \cite{radford2019language,10.5555/3495724.3495883,JMLR:v25:23-0870}, there has been an endless exploration of using language models to integrate multiple modules to achieve end-to-end optimization, such as SimpleTOD \cite{hosseini2020simple}, SOLOIST \cite{peng-etal-2021-soloist}, and UBAR \cite{yang2021ubar} to model multiple modules as sequence prediction processes to achieve end-to-end joint learning. After the release of large language models (LLM) such as ChatGPT \cite{chatgpt}, many studies focused on building task-oriented dialogue systems directly using LLM, such as SGP-TOD \cite{zhang-etal-2023-sgp}, InstructTODS \cite{chung-etal-2023-instructtods}, and AutoTOD \cite{xu-etal-2024-rethinking}, which use prompts to activate the ability of LLM to model task-oriented dialogue systems.

While TODS'core competency lies in their ability to process and generate natural languages, their practical application often depends on seamless integration with external services facilitated through APIs. The API acts as a bridge between the TODS and various back-end systems, databases, or third-party services, supporting the retrieval of necessary information or the execution of the requested action. For example, a travel reservation system can use APIs to query flight schedules, hotel availability, and prices from different providers. However, the lack of access to APIs makes the practical adoption of TODS challenging, and the dataset collected in this paper aims to build task-oriented dialogue systems with more widely accepted GUIs, freeing them from the security or cost constraints imposed by APIs.

\subsection{Multimodal Dialogue}

Compared with single-modal text dialogue, the development of multimodal dialogue is more in line with the actual needs. By mapping the data of various modes to a unified semantic space, the context association between different modes can be captured and utilized, so as to achieve more scenes, more functions and more accurate dialogue interaction. The most widely recognized multimodal dialogue task pertains to Image Grounded Conversations \cite{mostafazadeh-etal-2017-image,shuster-etal-2020-image} in open domains, where dialogue systems must engage with users around provided images and accurately respond to questions pertinent to the image content. As research progresses, the number of images incorporated into these dialogues has become increasingly abundant, posing greater challenges for the understanding of multiple images \cite{feng-etal-2023-mmdialog}. In recent years, the advancement of large multimodal models \cite{NEURIPS2023_9a6a435e,10.5555/3666122.3667638} has significantly propelled the progress of multimodal dialogue. By utilizing special tokens as triggers, various tasks such as Visual Grounding and Referring Expression Segmentation (RES) have also been gradually integrated into multimodal dialogue \cite{Lai_2024_CVPR,pmlr-v235-zhang24bu}, further broadening the scope of this field.

In the context of multimodal task-oriented dialogues, dialogue systems are often required to utilize images of items or products provided by users or retrieved from databases to assist users in accomplishing diverse tasks such as restaurant reservations, hotel selections, and shopping \cite{kottur-etal-2021-simmc}. Currently, research into leveraging GUI manipulation to access external databases for completing complex, multi-turn task-oriented dialogues remains an area ripe for exploration. The work most similar to ours is META-GUI \cite{sun-etal-2022-meta}, which accomplishes tasks across six domains—weather, calendar, search, taxis, restaurants, and hotels—by manipulating 11 different Android applications. In contrast, our work focuses on manipulating web pages, covering five domains including restaurants, hotels, tourist attractions, trains, and taxis. Moreover, the dataset we have collected contains complex dialogues spanning multiple domains, where the dialogue system needs to perform more challenging tasks such as domain switching.

\subsection{GUI Navigation}

By simulating human usage habits and operational logic, GUI Navigation is dedicated to building agents to automate web pages, desktops, and applications, enabling high-level natural language-driven automation \cite{li-etal-2020-mapping}. In recent years, the release of diverse datasets and benchmarks has significantly propelled the advancement of GUI navigation and automation. For instance, platforms such as WebShop \cite{NEURIPS2022_82ad13ec}, Mind2Web \cite{deng2023mind2web}, and WebArena \cite{zhou2023webarena} have focused on building automated agents within web scenarios, while OSWorld \cite{OSWorld} and WindowsAgentArena \cite{Bonatti2024WindowsAA} strive to achieve automated control over operating systems. Furthermore, there has been increasing exploration into mobile application automation, with benchmarks like MoTIF \cite{burns2022dataset}, AitW \cite{rawles2023androidinthewild}, AitZ \cite{zhang-etal-2024-android}, and GUI Odyssey \cite{lu2024gui} leading the charge. With the rapid development of large language models, research on GUI navigation has entered a phase of accelerated development, as evidenced by the exploration of GPT-3.5, ChatGPT, and GPT-4V for automated GUI manipulation \cite{10.1145/3544548.3580895,NEURIPS2023_7cc1005e,wen2023empowering}. To overcome the limitations of closed-source models, initiatives like Auto-GUI \cite{zhang-zhang-2024-look}, CogAgent \cite{10655402}, SeeClick \cite{cheng-etal-2024-seeclick}, UGround \cite{Gou2024NavigatingTD}, and ShowUI \cite{lin2024showui} have emerged, leveraging fine-tuned large multimodal models to enable automated GUI navigation.

In general, the core of GUI navigation tasks lies in understanding the functionality and layout of the GUI, locating target items and elements, and planning and executing tasks, with a focus on adhering to natural language instructions. In contrast, the datasets we collected emphasize the dialogue between users and agents, requiring the agent to possess the ability to understand the GUI and engage in proactive inquiries and conversations.

\section{MMWOZ}
In this section, we present a detailed process for constructing the MMWOZ dataset, which primarily involves the following steps: (1) Developing a web-style GUI based on the database content provided by MultiWOZ 2.3. (2) Designing an automated script to convert the dialogue state and system action annotation from MultiWOZ 2.3 into operation instructions for the GUI, and collecting snapshots of the web pages after executing these instructions. Additionally, we perform a detailed data analysis of MMWOZ and compare it with other relevant datasets.

\subsection{GUI Development}

The datasets such as MultiWOZ and CrossWOZ \cite{zhu-etal-2020-crosswoz} are collected using GUI-based annotation platforms in the form of the Wizard of Oz \cite{kelley1984iterative, wen-etal-2017-network}, where annotators simulate user and system interactions. However, these GUI platforms are specifically designed for data annotation purposes and differ significantly from actual graphical interfaces. Therefore, our initial task in creating MMWOZ is to develop a GUI that closely resembles real-world applications.

According to the authors' design in MultiWOZ 2.3, the agent assumes the role of a staff member at the Cambridge Town Information Center, assisting users with inquiries in 7 domains. In this paper, we focus on 5 domains: restaurant, hotel, attraction, train, and taxi. These domains are combined within a web-style GUI, comprising different panels as depicted in Figure \ref{GUI}. Our developed GUI encompasses a header bar that displays essential information, a menu bar for seamless navigation across domains, and a domain panel bar housing subpanels with distinct functions. For instance, as illustrated in Figure \ref{GUI}, the restaurant domain panel encompasses three subpanels: the finding subpanel for restaurant queries and listings, the information subpanel for displaying restaurant details, and the booking subpanel for making restaurant reservations. Additional GUI layout information for other domains can be found in Appendix \ref{appendix_gui}.

\subsection{Data Collection}

Due to the intricacy of GUI operations, most datasets pertaining to GUI operations, such as web navigation and app automation \cite{burns2022dataset, NEURIPS2022_82ad13ec, rawles2023androidinthewild, deng2023mind2web}, rely on manual annotation, which ensures the accuracy of operation instructions and imbues them with human-like logic. However, manual annotation incurs costs and is impractical for large-scale corpus. Fortunately, the MultiWOZ 2.3 dataset provides structured dialogue state and system action annotations, which can be leveraged to automatically generate GUI operation instructions. By associating this annotation information with elements on the web page using a script, the need for manual annotation is eliminated, thereby reducing costs. In MultiWOZ 2.3, the dialogue between the user and the agent is structured into a three-step process: first, the user provides information for a search; then, the agent identifies the relevant entity through recommendations; and finally, the agent completes the booking based on the provided reservation information. To follow this process, we design a three-step script that translates the dialogue state and system action annotations into operation instructions and captures web page snapshots.

\begin{breakablealgorithm}
\footnotesize
\caption{Pseudocode for collecting operation instructions and web page snapshots}
\label{alg1} 
\begin{algorithmic}[1]
    \REQUIRE dialogue state $S = [S_1, S_2, ..., S_t]$, system action $A = [A_1, A_2, ..., A_t]$, domain $D = [D_1, D_2, ..., D_t]$, operation actuator $\Omega$, snapshot generator $\Theta$.
    \STATE {operation instruction $P = [\ ]$, snapshot $N = [\ ]$, interface domain $\tilde{D} = \emptyset$, checked entity $\check{E} = [\ ]$.}
    \FOR {turn $i$ in $[1, 2, ..., t]$}
    \STATE{snapshot $n = call(\Theta)$, $N = N + [n]$, updated slot $S_{\delta}=S_i-S_{i-1}$}
    \FOR {domain $d$ in $D_i$}
    \IF {$d \neq \tilde{D}$}
    \STATE {operation $p = (click, (menu, d))$, $P = P + [p]$, $call(\Omega(p))$, snapshot $n = call(\Theta)$, $N = N + [n]$, $\tilde{D}=d$}
    \ENDIF
    \STATE{operation instruction $P_f = [\ ]$, target subpanel $r=$ finding subpanel}
    \FOR{slot $s$ in $S_{\delta}$}
    \IF{$s \in$ finding subpanel}
    \STATE{value $v = S_{\delta}(s)$}
    \IF{$s$ is click item}
    \STATE{operation $p = (click, (d, r, s, v))$}
    \ENDIF
    \IF{$s$ is input item}
    \STATE{operation $p = (input, (d, r, s), v)$}
    \ENDIF
    \STATE{$P_f = P_f + [p]$, $call(\Omega(p))$}
    \ENDIF
    \ENDFOR
    \IF{$P_f \neq [\ ]$}
    \STATE{target item $b=$ search button in finding subpanel, operation $p = (click, (d, r, b))$, $P_f = P_f + [p]$, $call(\Omega(p))$, $P = P + P_f$, snapshot $n = call(\Theta)$, $N = N + [n]$}
    \ENDIF
    \IF{$d \neq taxi$}
    \STATE{$S_{A_i}=$ mentioned slot in $A_i$, entity slot $S_e=$ \{name, id, phone, postcode, address\}}
    \IF{$S_{A_i} \cap S_e \neq \emptyset$}
    \STATE{entity $e=$ checked entity in $A_i$, $\check{E} = \check{E} + [e]$, operation $p = (click, (d, r, e))$, $P = P + [p]$, $call(\Omega(p))$, snapshot $n = call(\Theta)$, $N = N + [n]$}
    \ENDIF
    \IF{$d \neq attraction$ and $\check{E} \neq [\ ]$}
    \STATE{operation instruction $P_b = [\ ]$, target subpanel $r=$ booking subpanel}
    \FOR{slot $s$ in $S_{\delta}$}
    \IF{$s \in$ booking subpanel}
    \STATE{value $v = S_{\delta}(s)$, operation $p = (input, (d, r, s), v)$, $P_b = P_b + [p]$, $call(\Omega(p))$}
    \ENDIF
    \ENDFOR
    \STATE{$A_b=$ trigger action for booking in $A_i$}
    \IF{$A_b \neq \emptyset$}
    \STATE{target item $b=$ book button in booking subpanel, operation $p = (click, (d, r, b))$, $P_b = P_b + [p]$, $call(\Omega(p))$}
    \ENDIF
    \IF{$P_b \neq [\ ]$}
    \STATE{$P = P + P_b$, snapshot $n = call(\Theta)$, $N = N + [n]$}
    \ENDIF
    \ENDIF
    \ENDIF
    \ENDFOR
    \ENDFOR 
    \RETURN {$P$, $N$}
\end{algorithmic}
\end{breakablealgorithm}

The pseudocode of the script used to collect the data is shown in Algorithm \ref{alg1}, given the dialogue state $S$, system action $A$ and domain $D$ of a t-turn dialogue, using the designed web operation actuator $\Omega$ and snapshot generator $\Theta$, the collection process of operation instructions and snapshots is as follows: (1) For each domain mentioned in each turn of the dialogue (from lines 2 to 7), use the slot value information provided by the user to manipulate the elements in the finding subpanel and determine whether the $search$ button needs to be clicked (from lines 8 to 23). (2) For an entity whose $id$, $name$, $phone$, $postcode$, or $address$ is mentioned by the system action, open the information subpanel and booking subpanel corresponding to the entity (lines 24 to 28). (3) For the selected entity, input the booking information into its booking subpanel and click the $book$ button to complete the booking (lines 29 to 43).

\begin{figure}[!t]
    \centering
    \includegraphics[width=\linewidth]{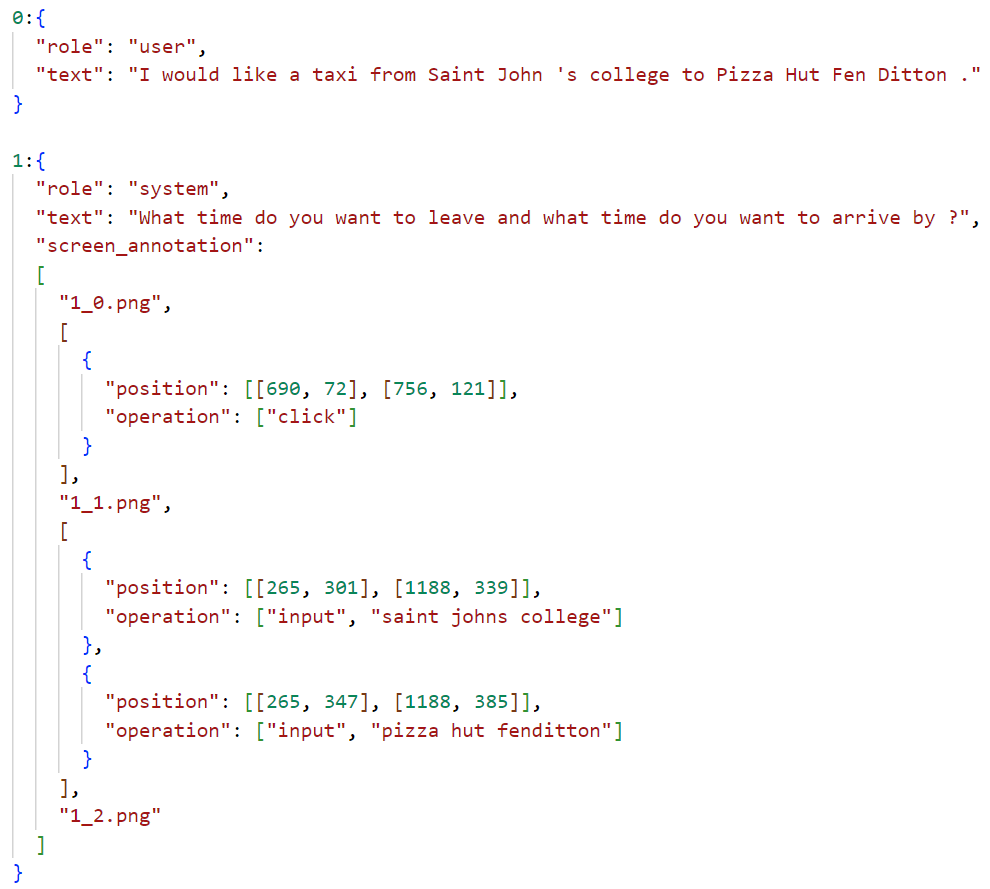}
    \caption{An example of how the data is organized in the MMWOZ dataset (From dialogue "SNG0073").}
    \label{annotation}
\end{figure}

In preparation for data collection, the snapshot generator utilizes Selenium\footnote{\url{https://www.selenium.dev/documentation/}} to control the Firefox\footnote{\url{https://www.mozilla.org/en-US/firefox/}} browser and capture screenshots. Meanwhile, the operation actuator can directly employ PyAutoGUI\footnote{\url{https://pyautogui.readthedocs.io/en/latest/}} to control the mouse and keyboard. Additionally, to enable the operation actuator to execute instructions composed of web page elements, the layout of the web page, including element coordinates, is obtained in advance. Upon collection, the MMWOZ dataset is organized as shown in Figure \ref{annotation}. Within the system role, the "screen\_annotation" item records a series of GUI operations along with corresponding snapshots. These GUI operations contain the coordinates of the manipulated web page elements and the content of the operations.

\begin{table*}[width=\textwidth,cols=7]
    \caption{Comparison of MMWOZ with existing task-oriented dialogue datasets (Training set).}
    \label{dataset_compare}
    \begin{tabular*}{\tblwidth}{@{ }LLLLLLL@{ }}
        \toprule
        \textbf{Metric} & \textbf{MultiWOZ} & \textbf{SGD} & \textbf{SIMMC 2.0} & \textbf{SpokenWOZ} & \textbf{META-GUI} & \textbf{MMWOZ}  \\
        \midrule
        Domains & 7 & 16 & 2 & 8 & 6 & 5 \\
        Slots & 30 & 214 & 16 & 36 & - & 30 \\
        Modalities & Text & Text & Text, image & Text, audio & Text, image & Text, image \\
        Interface & API & API & API & API & GUI & GUI \\
        Platform & - & - & - & - & \makecell[L]{Real-world,\\11 mobile APPs} & \makecell[L]{Simplified,\\5 web pages} \\
        Dialogues & 8,438 & 16,142 & 7,307 & 4,200 & 897 & 7,867 \\
        Turns & 115,424 & 329,964 & 38,127 & 149,126 & 3,692 & 109,558 \\
        \bottomrule
    \end{tabular*}
\end{table*}

\begin{figure}
    \centering
    \includegraphics[width=\linewidth]{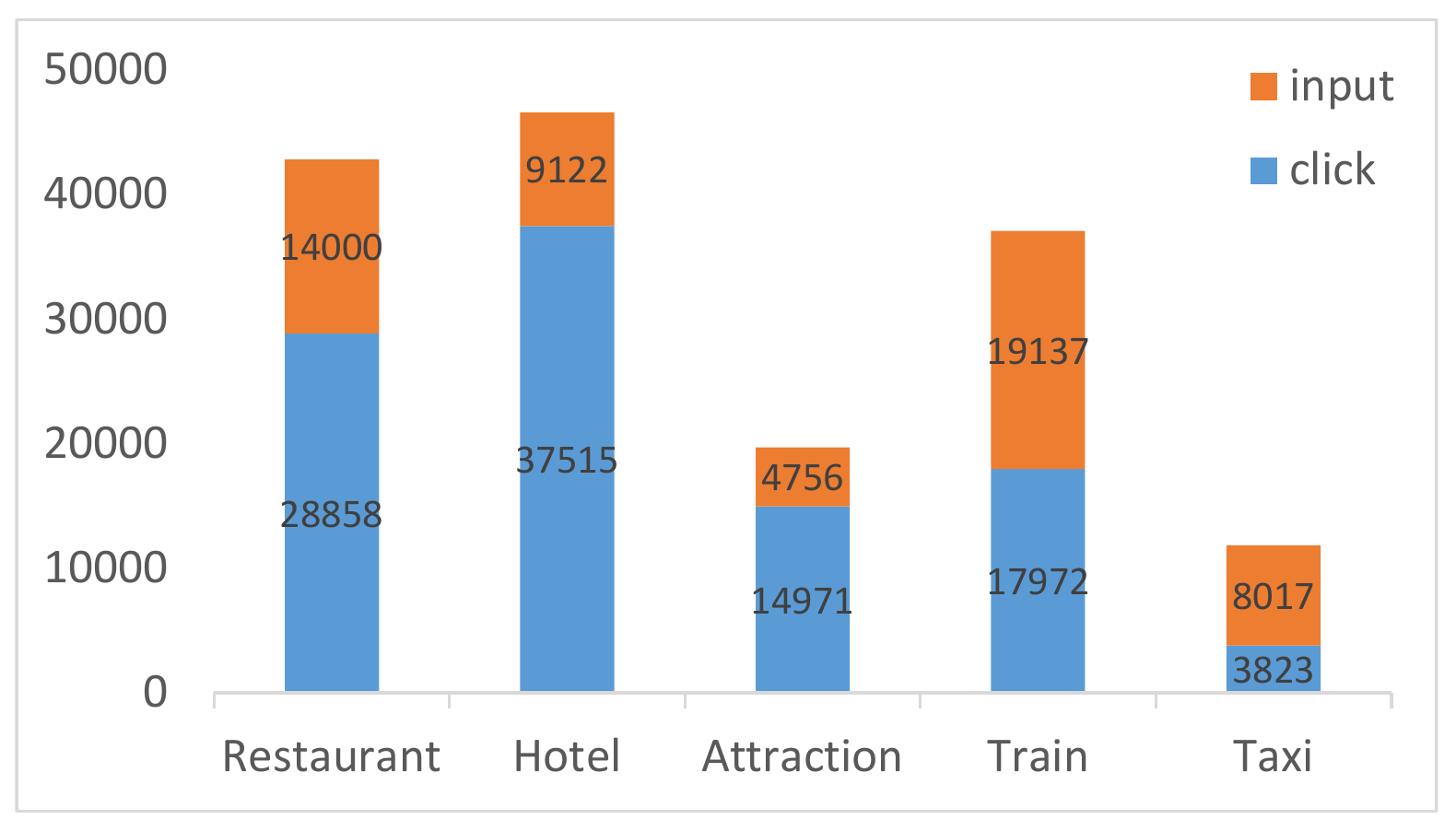}
    \caption{Distribution of operation types in different domains.}
    \label{domain_operation}
\end{figure}

\subsection{Data Analysis}

The MMWOZ dataset contains a total of 9,849 dialogues after filtering out samples that cannot be annotated by our script, and is split into a training, development and test set with 7,867, 990 and 992 dialogues, respectively. On average, each dialogue consists of 14.09 utterances, half of which are system utterances, with each system utterance corresponding to 2.16 web page snapshots and 2.28 operation instructions. In MMWOZ, operation instructions for the GUI are categorized as either "click" or "input" and their distribution across different domains is illustrated in Figure \ref{domain_operation}. Notably, the restaurant, hotel and attraction domains exhibit a higher proportion of "click" operations due to the presence of more categorical slots designed as checkboxes. In terms of operation frequency, the hotel domain has the highest number of operations, while the attraction and taxi domains have fewer operations compared to other domains, as they do not require detailed booking information.

The comparison between MMWOZ and existing task-oriented dialogue datasets is presented in Table \ref{dataset_compare}. Unlike traditional datasets that utilize APIs to access databases \cite{budzianowski-etal-2018-multiwoz, rastogi2020towards, kottur-etal-2021-simmc, si2023spokenwoz}, MMWOZ stands out by requiring the system agent to manipulate a web-style GUI to retrieve database information and respond to the user based on the displayed GUI information. The most similar dataset to MMWOZ is META-GUI \cite{sun-etal-2022-meta} dataset, which focuses on dialogue tasks involving 11 real-world Android mobile apps. However, MMWOZ primarily focuses on information-rich web-style GUI and includes a significantly higher number of dialogues with complex tasks compared to META-GUI.

\begin{figure*}
    \centering
    \includegraphics[width=\linewidth]{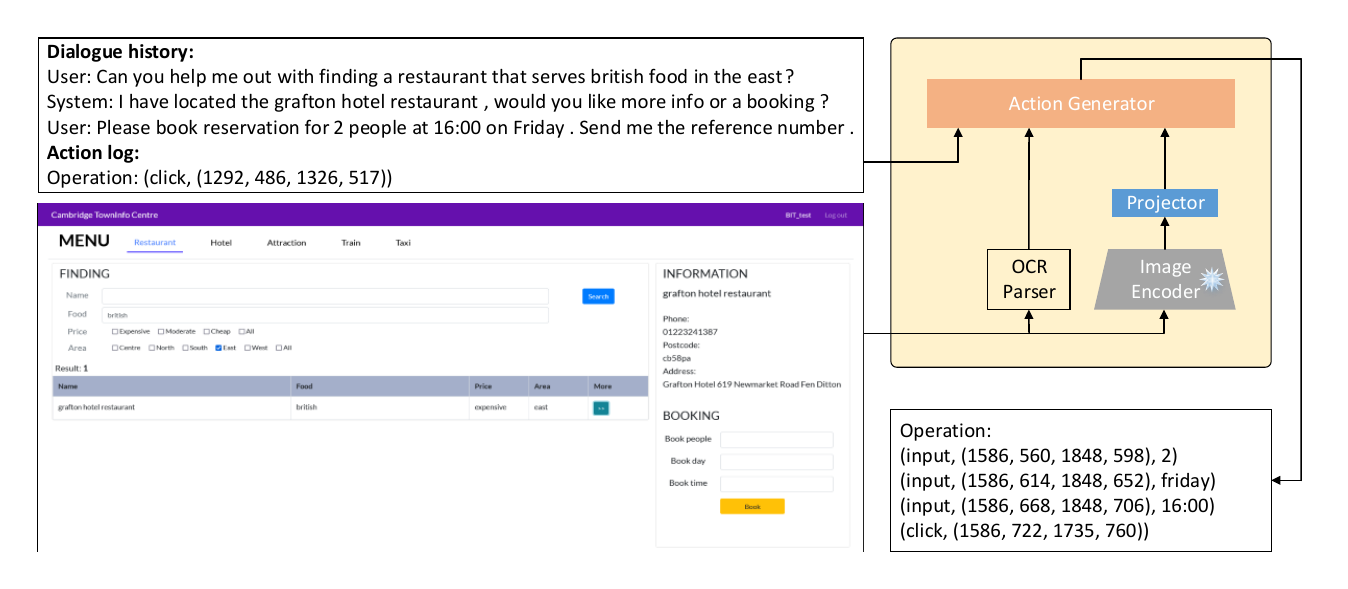}
    \caption{Model architecture of MATE. The dialogue history, action log, current web page snapshot, and the OCR result of the snapshot are fed to MATE to determine whether the next action is a sequence of operation instructions on the web page or a natural language response to the user.}
    \label{model}
\end{figure*}

\section{MATE}

The architecture of the MATE model is illustrated in Figure \ref{model}. To address the challenge posed by the complex tasks in the MMWOZ dataset, where the system role needs to operate the GUI multiple times to generate a reply. MATE is modeled at the action level, which needs to predict whether the next action is to generate the sequence of operation instructions for the GUI or the natural language response for the user. Specifically, the current web page snapshot is first processed by the OCR parser and image encoder to extract text information and obtain image features, respectively. Then, the action generator utilizes various inputs, including the dialogue history, the action log from the current turn, the OCR result, and the image features passed through the projector, to generate the next action.

\paragraph{OCR Parser}

The GUI provided by MMWOZ presents an information-rich web page, particularly containing a substantial amount of text information. The accuracy of the system's response to the user greatly relies on the correctness of this text information. To extract and capture the textual content from the web page snapshot, we employ Tesseract\footnote{\url{https://github.com/tesseract-ocr/tesseract}}, a widely-used open-source tool for Optical Character Recognition (OCR) parsing. Given a snapshot $N$ of a web page, using the OCR parser, the text information $N_{text}$ can be obtained:

\begin{equation}
N_{text}=OCRParser(N)
\end{equation}

\paragraph{Image Encoder}
In addition to the text information, the web page snapshot also contains the layout information of the elements in the web page. We use frozen ViT-B/16 of CLIP \cite{pmlr-v139-radford21a} as an image encoder to obtain the image feature $N_{image}$ in the snapshot:
\begin{equation}
N_{image}=ImageEncoder(N)
\end{equation}

\paragraph{Projector}
The image feature $N_{image}$ needs to undergo a dimensional transformation to ensure compatibility with the feature dimensions of other text information. This transformation is achieved using a trainable matrix $M$, which acts as a projector:
\begin{equation}
N^{'}_{image}=MN_{image}
\end{equation}

\paragraph{Action Generator}
To predict the next action, it is crucial to consider the dialogue history, which includes user goals and requirements, as well as the action log that contains completed actions. Therefore, we concatenate the dialogue history $H$, action log $L$, snapshot text information $N_{text}$ and snapshot image feature $N^{'}_{image}$ into an action generator with T5-small \cite{JMLR:v21:20-074} as the backbone, to generate the next action:
\begin{equation}
A_{next}=Generator(H, L, N_{text}, N^{'}_{image})
\end{equation}
where the next action $A_{next}$ is either a sequence of operation instructions aimed at retrieving further information from the current web page, or a natural language response to the user if the available information is deemed sufficient.

\section{Experiment Setup}

\begin{table*}[width=\textwidth,cols=10]
    \caption{Performance of the MATE model and its variants on the MMWOZ dataset. The second and third columns indicate the ablation settings for the dialogue history and action log included in the model's input.}
    \centering
    \begin{tabular*}{\tblwidth}{@{ }LLLLLLLLLL@{ }}
        \toprule
        \textbf{\multirow{3}{*}{Model}} & \textbf{\multirow{3}{*}{\makecell[L]{Dialogue\\History}}} & \textbf{\multirow{3}{*}{\makecell[L]{Action\\Log}}} & \textbf{\multirow{3}{*}{\makecell[L]{Action\\Type Acc.}}} & \multicolumn{4}{L}{\textbf{Operation}} & \multicolumn{2}{L}{\textbf{Response}}  \\
        \cmidrule(r){5-8} \cmidrule(r){9-10}
        & & & & \textbf{\makecell[L]{Location\\Acc.}} & \textbf{\makecell[L]{Command\\Acc.}} & \textbf{\makecell[L]{Snapshot\\Joint Acc.}} & \textbf{\makecell[L]{Turn\\Joint Acc.}} & \textbf{\makecell[L]{Entity\\Acc.}} & \textbf{\makecell[L]{BLEU\\Score}} \\
        \midrule
        \multirow{4}{*}{MATE$_{text}$}
        &  &  & 91.56 & 82.15 & 79.76 & 70.11 & 58.28 & 72.66 & 15.02 \\
        & \Checkmark &  & 91.80 & 85.52 & 84.43 & 75.06 & 64.79 & 68.56 & 14.29 \\
        &  & \Checkmark & 93.38 & 85.03 & 82.50 & 72.83 & 62.13 & 75.27 & 14.66 \\
        & \Checkmark & \Checkmark & \textbf{93.88} & \textbf{87.39} & \textbf{86.24} & 76.52 & 66.47 & \textbf{76.13} & \textbf{15.49} \\
        \hline
        \multirow{4}{*}{MATE$_{image}$} 
        &  &  & 83.94 & 75.79 & 73.50 & 64.16 & 50.21 & 12.97 & 9.42 \\
        & \Checkmark &  & 86.86 & 81.60 & 80.56 & 70.47 & 58.36 & 23.14 & 10.51 \\
        &  & \Checkmark & 92.51 & 83.45 & 80.69 & 68.21 & 55.60 & 13.11 & 10.88 \\
        & \Checkmark & \Checkmark & 93.86 & 85.04 & 83.85 & 72.42 & 61.47 & 29.75 & 15.21 \\
        \hline
        \multirow{4}{*}{MATE}
        &  &  & 91.18 & 83.03 & 80.63 & 70.46 & 58.49 & 71.75 & 15.04 \\
        & \Checkmark &  & 91.44 & 86.12 & 84.99 & 75.55 & 65.50 & 69.00 & 14.56 \\
        &  & \Checkmark & 93.14 & 85.27 & 82.65 & 73.02 & 62.22 & 74.45 & 14.68 \\
        & \Checkmark & \Checkmark & 93.47 & 87.31 & 86.05 & \textbf{76.67} & \textbf{66.72} & 73.87 & 14.41 \\
        \bottomrule
    \end{tabular*}
    \label{table_result}
\end{table*}

\subsection{Configuration}
We use the pre-trained T5-small to initialize the action generator in MATE. The image encoder employed is the frozen CLIP image encoder ViT-B/16, while the parameters in the projector are randomly initialized. During training, the model is trained for 10 epochs on the training set with a learning rate of 5e-4, utilizing Adam \cite{kingma2014adam} as the optimizer and a batch size of 16. In testing, the maximum output length is limited to 150. To investigate the significance of text information and image features in a web page snapshot, we also examine two other models within the MATE framework: MATE$_{text}$, which excludes image features, and MATE$_{image}$, which excludes OCR results. Both of these models follow the same training setup as MATE.

\subsection{Evaluation Metrics}
To evaluate the model's performance, we employ action type accuracy as a metric to determine whether the model correctly predicts whether the next action should involve operating the web page or responding to the user. In the case of operation instructions, we assess the accuracy of predicted coordinates and the entire operation using location accuracy and command accuracy for individual operations represented by ("click", coordinate) tuples or ("input", coordinate, value) triples. Additionally, we utilize snapshot joint accuracy and turn joint accuracy to verify the correctness of all operation instructions within each action or turn.

Regarding responses, apart from the BLEU score commonly used in previous literature \cite{budzianowski-vulic-2019-hello, hosseini2020simple, peng-etal-2021-soloist}, we also consider entity accuracy to assess whether the response includes relevant information such as address, phone number, postcode, reference number, and other associated entities. This provides a comprehensive evaluation of the model's ability to correctly incorporate entity-related information into its responses.

\section{Experiment Result}

\begin{table}[width=1.0\linewidth,cols=5]
\centering
\caption{The experimental results after including entity information coordinates in the response. \emph{Ref.} refers to the reference coordinate of the entity information from the web page snapshot.}
\label{table_analysis_ref}
\begin{tabular*}{\tblwidth}{@{ }LLLLL@{ }}
    \toprule
    \textbf{Model} & \textbf{Ref.} & \textbf{\makecell[L]{Turn\\Joint Acc.}} & \textbf{\makecell[L]{Entity\\Acc.}} & \textbf{\makecell[L]{BLEU\\Score}} \\
    \midrule
    \multirow{2}{*}{MATE$_{text}$} 
    & & 66.47 & 76.13 & 15.49 \\
    & \Checkmark & 65.91 & 75.41 & 15.78 \\
    \hline
    \multirow{2}{*}{MATE$_{image}$} 
    & & 61.47 & 29.75 & 15.21 \\
    & \Checkmark & 63.45 & 29.17 & 12.32 \\
    \hline
    \multirow{2}{*}{MATE}
    & & \textbf{66.72} & 73.87 & 14.41 \\
    & \Checkmark & 64.31 & \textbf{76.37} & \textbf{16.33} \\
    \bottomrule
\end{tabular*}
\end{table}

The performance of the MATE model and its variants on the MMWOZ dataset is shown in Table \ref{table_result}. It is evident that, in general, MATE$_{text}$ outperforms MATE in most metrics, except for snapshot joint accuracy and turn joint accuracy. For these two metrics, MATE$_{text}$ achieves an entity accuracy of 76.13\% and a BLEU score of 15.49. This outcome is reasonable because the input to the image encoder is a highly compressed image with a resolution of 224*224, making it challenging to retain text information. However, the web page layout information embedded in the image features aids MATE in achieving good performance in snapshot joint accuracy and turn joint accuracy. On the other hand, MATE$_{image}$ performs poorly in all metrics, particularly in entity accuracy, which relies on text information in snapshots, reaching only 29.75\%. These results indicate that a multimodal model utilizing low-resolution images (such as 224*224) faces difficulties when adapting to the MMWOZ dataset, as it requires the utilization of text information from snapshots to respond to users.

Based on the findings from the ablation setting in Table \ref{table_result}, it is evident that both dialogue history and action log play crucial roles in the performance of MATE. Removing the dialogue history from MATE's input leads to a decrease in turn joint accuracy from 66.72\% to 62.22\%. This outcome can be attributed to the fact that the dialogue history contains the user's intention and demands, which guide the system in operating the GUI effectively. Furthermore, the action log significantly affects the prediction accuracy of the next action type and the quality of the response. This is evident in the results, as removing the action log from MATE's input results in a decrease in action type accuracy to 91.44\% and entity accuracy to 69.00\%. Overall, these results highlight the critical role of both dialogue history and action log in MATE, as they contribute to accurate predictions and the generation of high-quality responses.

\begin{table*}[width=\textwidth,cols=16]
\caption{The results of domain transfer on the MMWOZ dataset. The table is divided into two parts, representing the model's performance in the source and target domains during zero-shot domain transfer. The performance metrics measured are turn joint accuracy (TJA), entity accuracy (EA), and BLEU score (BS).}
\label{domain_transfer}
\begin{tabular*}{\tblwidth}{@{ }LLLLLLLLLLLLLLLL@{ }}
    \toprule
    \textbf{\multirow{2}{*}{\makecell[L]{Evaluation on\\Source Domains}}} & \multicolumn{3}{L}{\textbf{Exclude Hotel}} & \multicolumn{3}{L}{\textbf{Exclude Train}} & \multicolumn{3}{L}{\textbf{Exclude Attraction}} & \multicolumn{3}{L}{\textbf{Exclude Restaurant}} & \multicolumn{3}{L}{\textbf{Exclude Taxi}} \\
    \cmidrule(r){2-4} \cmidrule(r){5-7} \cmidrule(r){8-10} \cmidrule(r){11-13} \cmidrule(r){14-16}
    & \textbf{TJA} & \textbf{EA} & \textbf{BS} & \textbf{TJA} & \textbf{EA} & \textbf{BS} & \textbf{TJA} & \textbf{EA} & \textbf{BS} & \textbf{TJA} & \textbf{EA} & \textbf{BS} & \textbf{TJA} & \textbf{EA} & \textbf{BS} \\
    \midrule
    MATE$_{text}$ & 65.07 & \textbf{75.28} & 15.46 & \textbf{65.14} & 76.57 & \textbf{14.55} & 60.89 & 71.47 & 13.86 & \textbf{61.60} & 72.45 & 13.02 & 63.63 & \textbf{70.14} & 15.41 \\
    MATE$_{image}$ & 61.31 & 25.49 & 12.81 & 61.12 & 23.58 & 11.83 & 58.96 & 19.18 & 12.61 & 58.06 & 28.64 & 11.99 & 59.15 & 29.78 & 13.40 \\
    MATE & \textbf{65.47} & 74.77 & \textbf{15.83} & 64.68 & \textbf{77.64} & 14.32 & \textbf{64.59} & \textbf{73.66} & \textbf{15.28} & 60.42 & \textbf{72.94} & \textbf{13.32} & \textbf{63.78} & 67.54 & \textbf{15.44} \\
    \bottomrule
    \toprule
    \textbf{\multirow{2}{*}{\makecell[L]{Evaluation on\\Target Domain}}} & \multicolumn{3}{L}{\textbf{Hotel}} & \multicolumn{3}{L}{\textbf{Train}} & \multicolumn{3}{L}{\textbf{Attraction}} & \multicolumn{3}{L}{\textbf{Restaurant}} & \multicolumn{3}{L}{\textbf{Taxi}} \\
    \cmidrule(r){2-4} \cmidrule(r){5-7} \cmidrule(r){8-10} \cmidrule(r){11-13} \cmidrule(r){14-16}
    & \textbf{TJA} & \textbf{EA} & \textbf{BS} & \textbf{TJA} & \textbf{EA} & \textbf{BS} & \textbf{TJA} & \textbf{EA} & \textbf{BS} & \textbf{TJA} & \textbf{EA} & \textbf{BS} & \textbf{TJA} & \textbf{EA} & \textbf{BS} \\
    \midrule
    MATE$_{text}$ & \textbf{13.59} & 58.44 & 8.15 & 9.35 & \textbf{55.56} & \textbf{5.43} & \textbf{7.69} & 47.06 & 10.67 & \textbf{20.00} & \textbf{55.95} & \textbf{10.63} & 15.20 & \textbf{4.08} & \textbf{6.89} \\
    MATE$_{image}$ & 5.23 & 0.00 & 6.59 & 0.93 & 0.00 & 4.18 & 0.00 & 5.88 & 6.23 & \textbf{20.00} & 0.00 & 6.62 & \textbf{33.60} & 0.00 & 5.40 \\
    MATE & 13.24 & \textbf{59.74} & \textbf{9.62} & \textbf{13.08} & 50.00 & 3.85 & \textbf{7.69} & \textbf{58.82} & \textbf{13.33} & 17.95 & \textbf{55.95} & 10.39 & 24.00 & 1.02 & 6.31 \\
    \bottomrule
\end{tabular*}
\end{table*}

\subsection{Reference Impact}

Considering the subpar performance of MATE in system responses, we introduce an enhancement based on the reference dialogue proposed in Shikra \cite{chen2023shikra}. This enhancement involves augmenting the entity information in the response with its corresponding reference coordinates from the snapshot. For instance, in Figure \ref{model}, the system utterance "I have located the grafton hotel restaurant" is predicted as "I have located the grafton hotel restaurant (35, 473, 588, 530)". The results presented in Table \ref{table_analysis_ref} demonstrate a significant improvement in the quality of responses generated by MATE through the inclusion of reference coordinates. The entity accuracy and BLEU score of MATE are improved to 76.37\% and 16.33, respectively. However, it is worth noting that the turn joint accuracy of MATE has dropped to 64.31\%. Moreover, both MATE$_{text}$ and MATE$_{image}$ exhibit noticeable drawbacks in their performance after the incorporation of reference coordinates. In summary, the inclusion of reference coordinates in the responses has shown promising improvements in the quality of MATE's outputs. However, it also introduces certain trade-offs, as evident from the decline in turn joint accuracy and the observed defects in MATE$_{text}$ and MATE$_{image}$. Therefore, the utilization of reference coordinates in responses should be approached cautiously, considering the potential risks involved.

\subsection{Domain Transfer}

\begin{table*}[width=\textwidth,cols=10]
\centering
\caption{The performance of the model after the web page layout changes. Interactive refers to the elements of a web page that can be manipulated (click or input), while non-interactive refers to the other elements that are not.}
\label{layout_adaptation}
\begin{tabular*}{\tblwidth}{@{ }LLLLLLLLLL@{ }}
    \toprule
    \textbf{\multirow{3}{*}{Model}} & \multicolumn{2}{L}{\textbf{Changed elements}} & \textbf{\multirow{3}{*}{\makecell[L]{Action\\Type Acc.}}} & \multicolumn{4}{L}{\textbf{Operation}} & \multicolumn{2}{L}{\textbf{Response}}  \\
    \cmidrule(r){2-3} \cmidrule(r){5-8} \cmidrule(r){9-10}
    & \textbf{Interactive} & \textbf{Non-interactive} & & \textbf{\makecell[L]{Location\\Acc.}} & \textbf{\makecell[L]{Command\\Acc.}} & \textbf{\makecell[L]{Snapshot\\Joint Acc.}} & \textbf{\makecell[L]{Turn\\Joint Acc.}} & \textbf{\makecell[L]{Entity\\Acc.}} & \textbf{\makecell[L]{BLEU\\Score}} \\
    \midrule
    \multirow{4}{*}{MATE$_{text}$} 
    &  &  & 93.88 & 87.39 & 86.24 & 76.52 & 66.47 & 76.13 & 15.49 \\
    & \Checkmark &  & 93.65 & 26.14 & 15.30 & 15.74 & 10.24 & 75.62 & 15.22 \\
    &  & \Checkmark & 93.99 & 87.35 & 86.20 & 76.20 & 66.12 & 70.81 & 15.05 \\
    & \Checkmark & \Checkmark & 93.63 & 25.91 & 15.04 & 15.32 & 10.05 & 69.63 & 14.63 \\
    \hline
     \multirow{4}{*}{MATE$_{image}$} 
    &  &  & 93.86 & 85.04 & 83.85 & 72.42 & 61.47 & 29.75 & 15.21 \\
    & \Checkmark &  & 93.73 & 24.84 & 13.78 & 13.45 & 8.24 & 29.88 & 15.09 \\
    &  & \Checkmark & 93.33 & 81.58 & 80.47 & 68.00 & 55.36 & 29.58 & 14.78 \\
    & \Checkmark & \Checkmark & 93.19 & 22.84 & 12.06 & 11.29 & 5.88 & 29.40 & 14.75 \\
    \hline
    \multirow{4}{*}{MATE}
    &  &  & 93.47 & 87.31 & 86.05 & 76.67 & 66.72 & 73.87 & 14.41 \\
    & \Checkmark &  & 93.49 & 27.41 & 17.31 & 16.56 & 10.22 & 72.92 & 14.39 \\
    &  & \Checkmark & 93.48 & 87.23 & 85.95 & 76.36 & 66.39 & 67.52 & 13.96 \\
    & \Checkmark & \Checkmark & 93.44 & 27.12 & 17.07 & 16.16 & 10.07 & 67.60 & 13.88 \\
    \bottomrule
\end{tabular*}
\end{table*}

In practical applications, as businesses undergo dynamic adjustments, new domains may need to be incorporated into existing dialogue systems. Therefore, it is essential for task-oriented dialogue systems built on the original GUI to possess the capability of transferring to these new domains. Specifically, when transferring the source model to a new domain, it is crucial for the model to not only understand how to operate the new GUI but also how to respond to user queries based on the content within the snapshot of the new GUI. Following the setting outlined in prior literature \cite{wu-etal-2019-transferable, yang2021ubar}, the domain transfer performance of the MATE model and its variants on the MMWOZ dataset is presented in Table \ref{domain_transfer}.

It is evident that the MATE model trained on the source domain data can still recognize entity information in the new GUI, as the names of entity information such as address, postcode, phone, and reference number are shared across domains. For instance, when considering the target domains of restaurant, hotel, attraction, and train, the entity accuracy exceeds 50\%. However, the model's ability to manipulate the new GUI is significantly reduced, with only 10\% to 20\% of the turn joint accuracy maintained. This highlights the complexity of transferring GUI operation instructions. Operating a new GUI requires the model to not only extract user requirements from their utterances but also locate the corresponding elements within the GUI. We believe that developing this capability necessitates the model's exposure to diverse GUI layouts in a large-scale corpus.

\subsection{Layout Adaptation}

The variation in display devices, such as changes in resolution, can result in adjustments to the layout of elements in the GUI. Consequently, the model's ability to adapt to these changes in element positioning becomes crucial for its practical application. In this paper, we introduce a categorization of GUI elements into two types: interactive elements, which can be manipulated, and non-interactive elements, which contain entity information. We then proceed to evaluate the model's adaptability to layout changes by altering the positioning of these two element types within the GUI.

The findings presented in Table \ref{layout_adaptation} demonstrate the challenges faced by MATE and its variants in adapting to changes in the layout of interactive elements. When the position of interactive elements is altered, both MATE and its variants exhibit a significant drop in turn joint accuracy, reaching approximately 10\%. This suggests that the model does not effectively learn the operational logic of interactive elements but rather relies on a direct association between coordinates and values. In contrast, for non-interactive elements, MATE and its variants demonstrate the ability to recognize and capture entity information even when their positions are modified. These models show less sensitivity to coordinate changes in non-interactive elements. One possible explanation for this outcome is that text is inherently contextual and remains consistent despite variations in coordinates. 

\subsection{Case Visualization}

\begin{figure*}
    \centering
    \includegraphics[width=\linewidth]{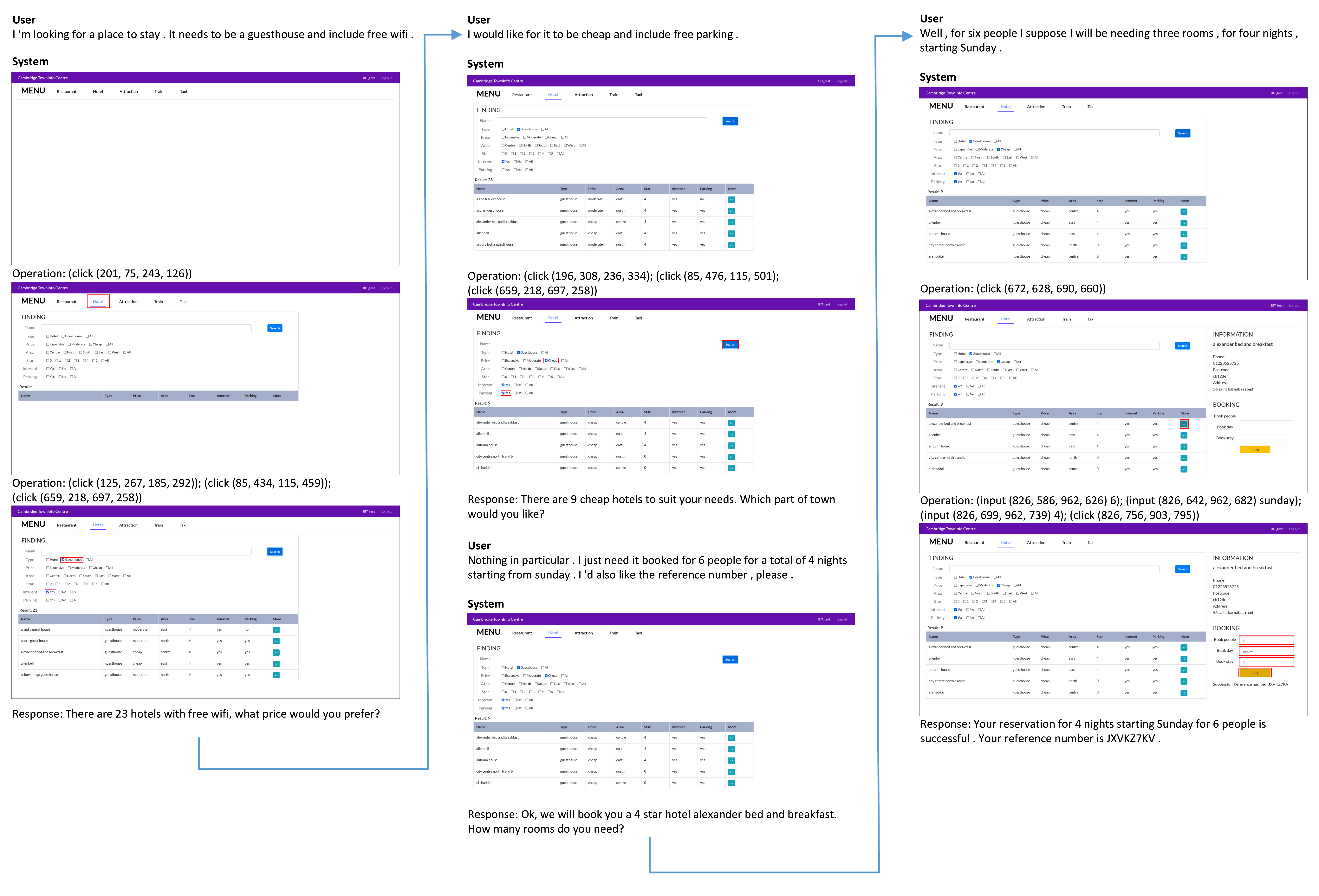}
    \caption{An example of GUI manipulation and dialogue response using MATE model on MMWOZ dataset, the coordinate positions have been marked in red boxes. (From dialogue "MUL0003").}
    \label{example}
\end{figure*}

To visualize the capability of the MATE model in manipulating a GUI for task-oriented dialogue, we manually captured screenshots and executed operations to obtain a detailed dialogue process of the model, as illustrated in Figure \ref{example}. Upon observing, it is evident that after extensive data training, the MATE model can manipulate the GUI by outputting precise operational instructions that include location coordinates. Furthermore, it leverages the information presented within the GUI to engage in dialogue with users.

\section{Conclusion}
In this paper, we introduce MMWOZ, a new multimodal task-oriented dialogue dataset. MMWOZ presents a unique challenge as the system not only engages in conversation with the user but also operates a web-style graphical user interface (GUI) to retrieve the required database information within the dialogue. To address this challenge, we propose a novel model called MATE, which serves as a baseline on the MMWOZ dataset. Furthermore, leveraging the constructed MMWOZ dataset and the proposed MATE model, we conduct a detailed analysis of the critical factors that need to be considered when building a practical and effective multimodal task-oriented dialogue system.

\appendix

\section{GUI Details}
\label{appendix_gui}
\begin{figure*}
    \centering
    \includegraphics[width=\linewidth]{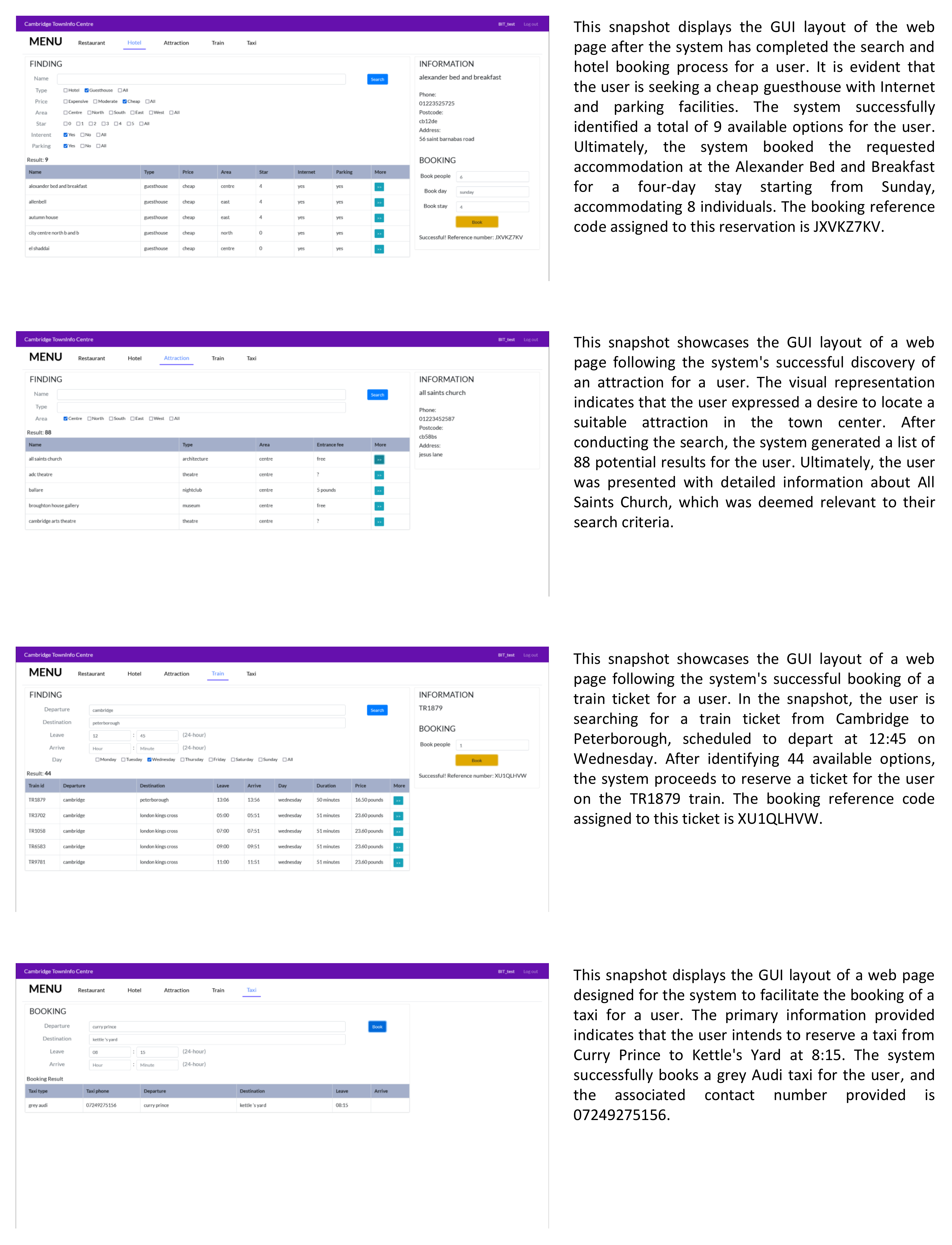}
    \caption{Examples of snapshots after using GUI to help users with tasks in different domains.}
    \label{GUI_domain}
\end{figure*}

In addition to the restaurant domain in Figure \ref{GUI}, the corresponding GUI for the other four domains are shown in Figure \ref{GUI_domain}.

\printcredits

\section*{Declaration of competing interest}
The authors declare that they have no known competing financial interests or personal relationships that could have appeared to influence the work reported in this paper.

\section*{Acknowledgment}
The work is supported by MIIT Program (CEIEC-2022-ZM02-0247), National Natural Science Foundation of China (No. U21B2009, 62172039 and 62276110).

\bibliographystyle{unsrt}
\bibliography{custom}

\end{document}